\documentclass[10pt,table]{article}

\usepackage[accepted]{rlj} 

\usepackage{graphicx}
\usepackage{amssymb}            
\usepackage{mathtools}          
\usepackage{mathrsfs}           
\mathtoolsset{showonlyrefs}     
\usepackage{graphicx}           
\usepackage{subcaption}         
\usepackage[space]{grffile}     
\usepackage{url}                
\usepackage{wrapfig}            
\usepackage{lipsum}             
\usepackage{cleveref}           
\usepackage{amsfonts}           
\usepackage{bbm}                

\usepackage{amsmath}
\DeclareMathOperator\supp{supp}

\newcommand{\Beta}{\text{Beta}}

\title{Shaping Laser Pulses with Reinforcement Learning}

\setrunningtitle{Shaping Laser Pulses with Reinforcement Learning}

\author{
Francesco Capuano\textsuperscript{1},
Davorin Peceli\textsuperscript{2},
Gabriele Tiboni\textsuperscript{3,4}
}

\emails{
francesco.capuano@ens-paris-saclay.fr, \\
davorin.peceli@eli-beams.eu, \\
gabriele.tiboni@uni-wuerzburg.de
}

\affiliations{
$^{1}$\textbf{École Normale Supérieure Paris-Saclay, Gif-sur-Yvette, France} \\
$^{2}$\textbf{Extreme Light Infrastructure-ELI Beamlines, ELI Beamlines, Dolní Břežany, Czechia }\\
$^{3}$\textbf{Julius-Maximilians-Universität Würzburg, Würzburg, Germany}\\
$^{4}$\textbf{Technische Universität Darmstadt, Darmstadt, Germany}
}


\summary{
High Power Laser (HPL) systems operate in the attoseconds regime---the shortest timescale ever created by humanity.
HPL systems are instrumental in high-energy physics, leveraging ultra-short impulse durations to yield extremely high intensities, which are essential for both practical applications and theoretical advancements in light-matter interactions. 
Traditionally, the parameters regulating HPL optical performance have been manually tuned by human experts, or optimized using black-box methods that can be computationally demanding.
Critically, black box methods rely on stationarity assumptions overlooking complex dynamics in high-energy physics and day-to-day changes in real-world experimental settings, and thus need to be often restarted. Deep Reinforcement Learning (DRL) offers a promising alternative by enabling sequential decision making in non-static settings.
This work explores the feasibility of applying DRL to HPL systems, extending the current research by (1) learning a control policy relying solely on non-destructive image observations obtained from readily available diagnostic devices, and (2) retaining performance when the underlying dynamics vary.
We evaluate our method across various test dynamics, and observe that DRL effectively enables cross-domain adaptability, coping with dynamics' fluctuations while achieving 90\% of the target intensity in test environments.
}

\contribution{We demonstrate the benefits of using Deep Reinforcement Learning to optimize High Power Laser systems over currently dominant approaches based on gradient-free optimization.}
{Prior works on laser optimization focused on black-box optimization techniques which assume stationarity, require costly real-world function evaluations, and can endanger the system at test-time.}

\contribution{We learn a control policy directly from images, which are made available via widespread diagnostics devices.}
{Instead of relying on noisy and lengthy processes to obtain structured representations of the system's state, we leverage unstructured observations coming from diagnostic devices as inputs for the control policy.}

\contribution{We train control policies entirely in simulation and successfully transfer them across unknown, varying dynamics, showing robustness to different parametrizations.}
{Transferring policies is hindered by the discrepancies across different domains, and Domain Randomization, a promising technique widely explored in then field of robot learing, can be leveraged to ensure robustness.}

\begin{document}

\makeCover  
\maketitle  

\begin{abstract}
High Power Laser (HPL) systems operate in the attoseconds regime---the shortest timescale ever created by humanity.
HPL systems are instrumental in high-energy physics, leveraging ultra-short impulse durations to yield extremely high intensities, which are essential for both practical applications and theoretical advancements in light-matter interactions. 
Traditionally, the parameters regulating HPL optical performance have been manually tuned by human experts, or optimized using black-box methods that can be computationally demanding.
Critically, black box methods rely on stationarity assumptions overlooking complex dynamics in high-energy physics and day-to-day changes in real-world experimental settings, and thus need to be often restarted. Deep Reinforcement Learning (DRL) offers a promising alternative by enabling sequential decision making in non-static settings.
This work explores the feasibility of applying DRL to HPL systems, extending the current research by (1) learning a control policy relying solely on non-destructive image observations obtained from readily available diagnostic devices, and (2) retaining performance when the underlying dynamics vary.
We evaluate our method across various test dynamics, and observe that DRL effectively enables cross-domain adaptability, coping with dynamics' fluctuations while achieving 90\% of the target intensity in test environments.
\end{abstract}

\section{Introduction}
\label{sec:introduction}
Ultra-fast light-matter interactions find applications in both theoretical and experimental physics. The extremely high intensities---in the order of petawatts---that can be attained with modern-day High Power Laser (HPL) systems enable a variety of use cases in light-matter interactions and charged-particles acceleration.
Extreme intensities are typically achieved by focusing high-energy laser pulses onto spatial targets for ultra-short durations---down to attoseconds. As a result, ultra-short (attoseconds) laser pulses represent the shortest events ever created by humanity~\citep{gaumnitz2017streaking}.

Over the course of 2022 and 2023, four separate experiments at the Lawrence Livermore National Laboratory (LLNL)-National Ignition Facility (USA) employed HPL systems to achieve nuclear fusion ignition~\citep{abu2024achievement}. 
In their experiments, the scientists at the LLNL used 192 HPL beams to achieve nuclear fusion ignition in a laboratory setting, and went on demonstrating larger-than-unity energy gains, achieving energy-positive results in nuclear fusion. 
HPL systems also have applications in radiation-based cancer therapy, as they can be used to produce beams of high-energy charged particles, which interact with malignant cells and thus yield radio-therapeutic outcomes~\citep{grittani2020device}. 
Lastly, HPL systems enable the controlled study of the interaction between extremely intense beams of light and various materials, providing valuable insights to numerous scientific communities, including theoretical, plasma, and laser physicists.

HPL systems' performance heavily depends on environmental conditions, and on numerous parameters. For instance, HPL systems are typically operated in remote areas or meters underground to mitigate road-induced vibrations that might cause misalignment in the optics. 
Further, HPL systems are run in environmentally controlled facilities (\textit{cleanrooms}), to prevent airborne particles from sedimenting on the optical gear. 
Parameters-wise, \textit{dispersion coefficients} play a central role, as they physically determine the phase shifts imposed on the different frequencies of the light beam. 
In turn, this leads to shorter laser pulses and intensity gains when the applied phase induces constructive interferences between frequencies, whereas destructive interference results in longer pulses and intensity losses~\citep{paschotta2008field}.

Traditionally, laser parameters have been optimized using extensive 1D searches over the entire range of possible values, repeating independent searches for all parameters. 
More recently, black-box numerical methods such as Evolution Strategies (ES) and Bayesian Optimization (BO) have been studied~\citep{loughran2023automated, shalloo2020automation, capuano2022laser, arteaga2014supercontinuum}. 
These black-box methods can be computationally demanding, as they require being implemented on real-world laser systems, resulting in costly laser-bursts to perform each function evaluation. 
Further, black-box methods rely on stationarity assumptions overlooking transient and complex non-linear system dynamics, hindering both (1) the transfer of solutions across simulations and real-world HPL systems and (2) efficient adaptation to ever evolving experimental conditions.
Lastly, their safe implementation on real-world hardware can be challenging, as the erratic exploration of the parameter space can compromise system safety~\citep{capuano2023temporl}.

\begin{figure}
    \centering
    \includegraphics[width=\linewidth]{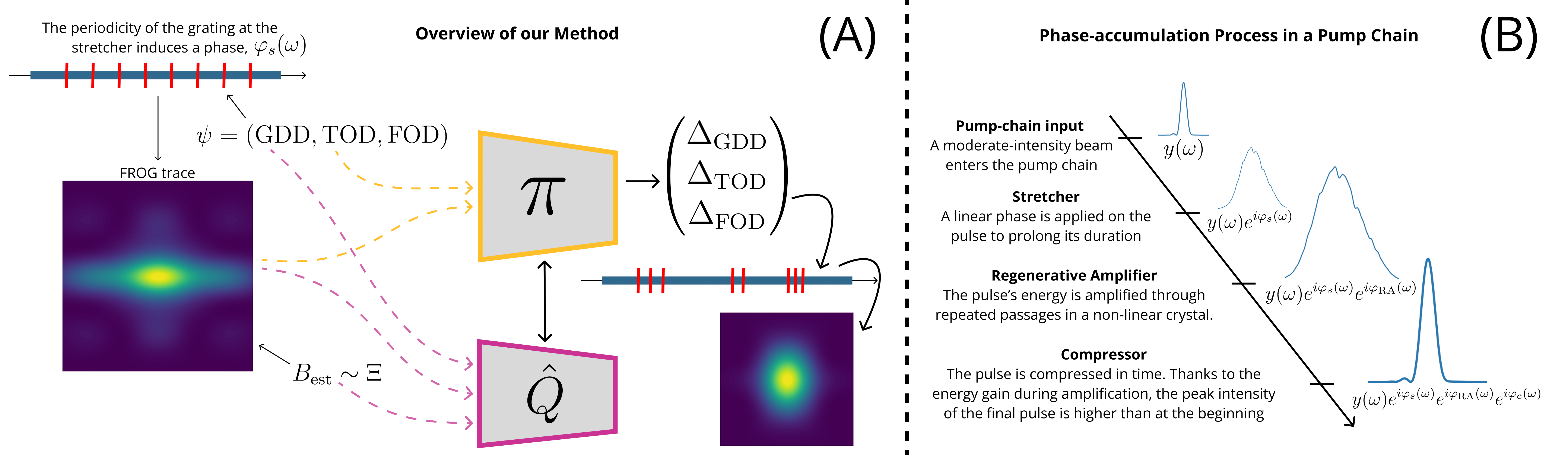}
    \caption{(A) Schematic representation of the RL pipeline for pulse shaping in HPL systems. The model processes images to produce phase corrections, leading to shorter pulse durations, by maximizing intensity. To improve on robustness, during training the agent faces \textit{a distribution} of dynamics rather than a single parametrization. (B) Illustration of the process of linear and non-linear phase accumulation taking place along the pump-chain of HPL systems. By opportunely controlling the phase imposed at the stretcher, one can benefit from both energy and duration gains, for maximal peak intensity.}
    \label{fig:figure1_and_cpa}
\end{figure}

This work investigates the safe application of DRL to HPL systems for temporal profile shaping via autonomous, bounded control of \textit{dispersion coefficients}. 
In particular, we present an application of DRL to intensity maximization through pulse duration minimization. 
We leverage an openly-available simulator~\citep{capuano2023temporl} of a component of the world's most powerful laser system~\citep{batysta2014pulse}, and learn an adaptive control policy capable of safely tuning the dispersion coefficients for intensity maximization. 
In our work, we simulate different experimental conditions by arbitrarily randomizing parameters of our simulator, and use randomization over the laser system dynamics to induce robustness to changes in experimental settings~\citep{tiboni2023domain}.
As parameters of HPL systems can typically only be estimated and vary over time, robustness is paramount for an effective application of our method. To further improve on this and pave the way towards real-world applications of RL to HPL systems, we exclusively process unstructured observations in the form of images (\textit{FROG traces}) readily available through diagnostic devices. 

Our contributions can be summarized as follows:
\begin{itemize}
    \item We present an \textbf{application of DRL to the rich and complex domain of experimental laser physics}, demonstrating its suitability in handling the non-stationary and transient non-linear dynamics of HPL systems---challenges often overlooked by prevailing black-box approaches.
    \item \textbf{We learn a control policy from single-channel images} readily available in most experimental settings, using them as a proxy for otherwise destructive pulse intensity measurements. Using images directly eliminates the need for quantum-destructive measurements of charged particles energy or noisy temporal pulse reconstruction, and provides feedback signal leveraging existing experimental hardware, making our method applicable in real-world settings.
    \item We \textbf{train control policies entirely in simulation and successfully transfer them} across different environments, ensuring adaptability to (1) inaccuracies in parameter estimation and (2) evolving experimental conditions. Randomizing also helps mitigate the impact of under-modeling in simulation.
\end{itemize}

We open-source environment and training code at \url{https://github.com/fracapuano/RLaser}.

\section{Background \& Related Work}
\label{sec:background}
\subsection{Optimizing Laser Systems}
Traditionally, HPL systems' parameters have been optimized using independent 1D grid-searches over all the considered dimensions. 
While straightforward, this approach naively overlooks the joint effect varying multiple parameters simultaneously can have on the system. 
More recently, Evolution Strategies (ES)~\citep{baumert1997femtosecond, arteaga2014supercontinuum, woodward2016towards}, and Bayesian Optimization (BO)~\citep{loughran2023automated, shalloo2020automation, capuano2022laser, anjum2024high} have been proposed to optimize HPL performance. 
Differently from grid-search, ES and BO do take into account the joint effect of different parameters on the system, and proved effective in real-world experiments~\citep{shalloo2020automation}. 
However, black-box methods tend to be computationally demanding in the number of functions evaluations---in practice, real-world laser bursts. 
While BO is significantly more sample efficient than ES~\citep{capuano2022laser}, both ES and BO typically do not provide guarantees regarding the stability of the control configuration to changes in the environment.
That is, for any changes in the experimental condition one could need to re-optimize the system from scratch, just as human experts would do. 
Indeed, black-box algorithms are used to optimize an unknown objective function, and typically rely on stationarity assumptions within experimental conditions, thus overlooking the transient and complex dynamics characteristic of high-intensity phase accumulation processes in non-linear crystals.
In addition---differently from grid search---the safe implementation of black-box methods on real-world hardware can be challenging, as gains in sample efficiency might trade-off with erratic exploration of the parameter space~\citep{capuano2023temporl}, endangering system's safety.
Lastly, being inherently oriented towards finding \textit{point} solutions---control \textit{parameters} rather than control \textit{strategies}---black-box methods cannot robustly exploit approximate \textit{simulations} of \textit{real-world} HPL systems, as the objective function's landscape may differ significantly across the \textit{reality gap}~\citep{tobin2017domain}.

To allow for a more adaptive control of laser systems, recent works have investigated the application of Reinforcement Learning (RL) to HPL systems~\citep{kuprikov2022deep, rakhmatulin2024reinforcement, mareev2023self, capuano2023temporl}. 
\citet{mareev2023self} investigated the application of DRL to maintain a laser beam focused on a solid target, shifting away as a consequence of high-energy light-matter interactions and thus requiring constant target-position adjustment. 
\cite{rakhmatulin2024reinforcement} investigated the application of RL to the problem of optics alignment in laser systems, controlling the position of mirrors via real-time camera feedback. 
While both target location and mirror alignment have a significant impact on the final intensity conveyed by the beam, neither directly shapes the temporal profile of laser pulses, and thus the final peak intensity. 
\cite{kuprikov2022deep} learned a controller to adaptively adjust the power supplied to the laser, and the filters used to temporally shape the output, thus directly impacting peak intensity. 
However, the authors considered the problem of ensuring highly-similar pulses between multiple laser bursts, by learning to mode-lock the system, rather than shaping the individual pulse. 
\cite{capuano2023temporl} studies the problem of learning a controller for pulse shaping, by directly tuning the dispersion coefficients, thus ensuring a closer loop between control parameters and peak intensity. 
However, \cite{capuano2023temporl} overlook several practical aspects associated with deploying control policies to real world laser systems, such as the necessity of coping with possibly imprecise estimates of the experimental conditions, and the need to adapt to the non-stationary of the experimental environment.
Unlike previous attempts at temporal pulse shaping, we work backwards from real-world deployment requirements, extending the current research by learning a robust control policy for the dispersion coefficients that is (1) machine-safe to deploy, (2) inherently adaptive and (3) uses \textit{FROG traces}, images readily available in most HPL diagnostic systems.
Accordingly, our method is not directly comparable to black-box methods such as BO or ES because (1) black-box methods are not suited to process unstructured observations such as images, and (2) black-box methods suffer from significant limitations in transferring point solutions across the reality gap.
Lastly, we note that attempting to use the optimal control parameters obtained with black-box methods \textit{in simulation} as an initialization for efficient \textit{real-world} optimization fundamentally depends on access to high-fidelity simulations, which remain largely unavailable for most HPL systems.

\subsection{Shaping Laser Pulses}
The optimization of laser pulse shape and duration is a critical challenge in HPL systems, particularly for applications in laser-plasma acceleration, high-intensity laser-matter interactions, and inertial confinement fusion. 
Furthermore, the precise control of pulse shape directly influences the peak intensity, energy deposition efficiency, and nonlinear optical effects encountered during laser propagation. In applications of HPL systems to charged particle acceleration~\citep{grittani2020device}, directly measuring the particles' beam energy is a quantum-destructive process---charged particles lose their energy when an experimental energy probe interacts with them.
However, proxying particles' beam energy with pulse's peak intensity, HPL systems can be optimized using the peak intensity \( I^* \) produced. 
At iso-energy, intensity maximization takes place by minimizing the pulse duration, measured by its full-width half-maximum (FWHM) value---the value \( \vert t_l - t_r \vert: I(t_l)=I(t_r)= \frac 12 I^* \). Ultra-short pulses' duration is typically inferred from \emph{frequency-resolved optical gating} (FROG) traces~\citep{trebino1993using}, for the scope of this work considered as single-channel \textit{images} visualizing the spectral phase accumulated by a pulse. 
Thus, black-box methods and 1D-grid search are fundamentally ill-posed to use these proxies of particle beam's energy as input, while DRL can instead fully leverage the advancements made in Deep Learning to handle unstructured data formats as control inputs~\citep{mnih2013playing}.

In practice, HPL systems rely on the energy transfer from a high-power primary \textit{pump} laser beam to a secondary \textit{seed} laser beam. The spectral and temporal characteristics of the pump laser determine much of the achievable pulse intensity. Critically, for the sake of intensity gains in the seed laser, the pump laser is usually run through an amplification chain introducing both linear and nonlinear phase distortions. As phase regulates how the spectral intensity overlays in the time domain~\citep{paschotta2008field}, it must be carefully controlled to achieve efficient amplification at the pump and seed level. 
Typically, pump chains follow a Chirped Pulse Amplification (CPA) scheme. Figure~\ref{fig:figure1_and_cpa} illustrates the CPA process, where the initial pump pulse is (1) stretched in time to avoid nonlinear effects and damage to the earlier stages of the pump chain due to high intensities (2) amplified via regenerative and multipass amplifiers, and (3) re-compressed in time to achieve high peak intensity.

Unlike the amplification and compression stages, the process of pulse stretching can typically be controlled externally from laser specialists, varying the dispersion coefficients of the phase applied on the pump laser. The spectral phase of a laser beam \( \varphi(\omega) \) is typically modeled using a Taylor expansion around the central angular frequency of the pulse \( \omega_0 \), yielding \(
\varphi(\omega) = \sum_{k=0}^{\infty} \frac 1{k!}\frac{\partial^k \varphi}{\partial \omega^k} (\omega - \omega_0)^k \).
The first two terms in the polynomial expansion---\( \varphi(\omega) \) and \( \varphi^\prime(\omega)(\omega - \omega_0) \)---do not directly influence the shape of the pulse in the temporal domain. Conversely, second-order (\textit{group-delay dispersion}, GDD), third-order (\textit{third-order dispersion}, TOD) and fourth-order (\textit{fourth-order dispersion}, FOD) derivatives---jointly referred to in this work with \( \psi = (\text{GDD}, \text{TOD}, \text{FOD}) \in \Psi \)---do influence the resulting temporal profile. 
By opportunely tuning \( \psi \), laser specialists are able to control the temporal profile of ultra-short laser pulses. Physically, control over $\psi$ is achieved using a Chirped Fiber Bragg Grating (CFBG), consisting of an optical fiber whose grating is adjusted inducing a temperature gradient at its extremes. Consequently, it is crucial to carefully regulate the relative temperature variations to avoid demanding abrupt control adjustments over short time intervals, which could damage the fiber.

In the context of laser optimization, one might want to maximize the intensity conveyed by a laser pulse by minimizing its duration, i.e. performing \textit{temporal shaping} by controlling \( \psi \). Typically, highly trained human experts spend hours carefully varying \( \psi \) in the real world, leveraging a mix of past experience and personal expertise at the task. The shortest time duration attainable by a laser pulse is typically referred to as Transform Limited (TL), and corresponds to perfect overlay of all the different spectral components of intensity in time---as such, it has an accumulated phase equal to \( \varphi^*(\omega)=0 \). Critically, the amplification step in CPA introduces nonlinear phase components. If this was not the case, then one could retrieve TL pulses by simply applying a phase at the stretcher level that is opposite to the one defined at the compressor's, \( \varphi_s(\omega) = - \varphi_c(\omega) \). However, the non-linearity induced by the amplification step calls for a more sophisticated control over, \( \varphi_c(\omega) \). This difficulty arises from the need to balance non-linear effects in the phase accumulation process and non-stationary experimental conditions, while adhering to a sequential control approach that ensures machine safety by limiting abrupt changes in control parameters~\citep{capuano2023temporl}.

\subsection{Sim-to-real}
Even the most sample efficient of the numerical algorithms typically considered for pulse shaping varying dispersion coefficients can require hundreads of samples~\citep{capuano2022laser}, corresponding to just as many real-world laser bursts~\citep{shalloo2020automation}. Such computational demands are hard to meet in real-world systems, and are especially more troubling if one considers the instability of the solution found with respect to changes in the experimental setting. Further, BO can endanger the system by applying abrupt controls at initialization.

We can mitigate the need for expensive real-world data samples by leveraging simulated versions of the phase accumulation process, where we can easily accommodate for large number of samples, as well as safe exploration of the dispersion coefficients space, \( \Psi \). While typically not accurate enough to directly transfer point-solutions \( \psi^* \) from simulations to the real world, simulators can be used to train control policies for different environments. The problem of transferring control policies across domains is a well-studied problem in applications of RL for robotics, and the community has extensively investigated approaches to crossing the \emph{reality gap}~\citep{tobin2017domain, valassakis2020crossing}. Considering this last point, we argue the HPL setting closely resembles the challenges the community faces when transferring policies across environments. 

Transferring a control policy across diverse environments can be achieved (1) reducing the discrepancy between them~\citep{zhu2017fast} and/or (2) applying parameter randomization to improve on the robustness of the policy~\citep{peng2018sim}.
As (1) typically requires significant modeling efforts, in this work we decide to focus on (2).
One widely adopted sim-to-real method is Domain Randomization (DR), which involves varying simulator parameters within a predefined distribution during training~\citep{valassakis2020crossing} to incentivize generalization over said parameters. DR introduces additional sources of stochasticity into the environment dynamics, making policies more robust but with the potential drawback of sub-optimality and over-regularization~\citep{margolis2024rapid}.

Although having proved effective on robotics tasks~\citep{antonova2017reinforcement}, DR suffers from the key limitation of needing to extensively tune the distributions used in training. Automated approaches to DR propose adaptive distribution refinement over training, e.g.~by leveraging a limited set of real-world data~\citep{tiboniadrbenchmark,tiboni2023dropo}, or based on the policy's performance under a given set of dynamics parameters~\citep{akkaya2019solving}. While effective for dexterous manipulation,~\citet{akkaya2019solving} has been observed to be sample inefficient, as it biases the policy towards learning dynamics sampled from the boundaries of the current distribution~\citep{tiboni2023domain}. A more principled approach to automated DR has been recently introduced by~\citet{tiboni2023domain}, where the authors follow the principle of maximum entropy~\citep{jaynes1957information} to resolve the ambiguity of defining DR distributions. In particular, the authors train adaptive control policies with progressively more diverse dynamics while satisfying an arbitrary performance lower bound.
Notably, the domain randomization approaches in~\citet{akkaya2019solving,tiboni2023domain} employ history-based policies to promote implicit meta-learning strategies at test time---i.e., on-line system identification.

\section{Method}
\label{sec:method}
\subsection{MDPs for Intensity Maximization}\label{sec:mdp}
In~\cite{capuano2023temporl}, the authors formulate pulse shaping as a control problem in a Markov Decision Process (MDP), \( \mathcal{M} \). In this work, we extend the MPD formulation to the case where the environment dynamics are influenced by an unobserved latent variable, leading to a \textit{Latent MDP} (LMDP)~\citep{chen2021understanding}, denoted as \(\mathcal{M}_\xi = \{\mathcal{S}, \mathcal{A}, \mathbb{P}_\xi, r, \rho, \gamma\}\). 
Here, \( \xi \) is a realization of a latent random vector \( \Xi \), such that \(\xi \sim \Xi : \supp(\Xi) \subseteq \mathbb{R}^{|\xi|}\), parametrizing the transition dynamics \( \mathbb{P}_\xi \). Crucially, the agent does not directly observe \( \xi \) in the test environment (i.e.~the real world).
Conversely, we assume that parameters \( \xi \) may be accessed when training in simulation.
We argue the LMDP framework is particularly well-suited for pulse shaping in a non-stationary setting due to the presence of hidden variations in the system's dynamics. In practical scenarios, an agent must adapt to an unknown experimental condition which can be modeled as \( \xi \), while iteratively refining its control \( \psi \). As \( \psi \) is physically translated into temperature gradients applied to an optical fiber, the choice of \( \psi_t \) must account for past applied controls, particularly \( \psi_{t-1} \), to prevent excessive one-step temperature variations, which may endanger the fiber. Moreover, the day-to-day fluctuations in HPL systems can be captured through \( \Xi\), modeling the inherent non-stationarity of experimental conditions. Further, by incorporating a distribution over the starting condition of the system, \( \psi_0 \sim \rho \), the pulse shaping problem's sequential nature becomes evident---starting from a randomly sampled experimental condition, the agent must iteratively apply controls \(\psi\) while dealing with incomplete knowledge of the system dynamics.
Inspired by the domain randomization and meta-learning literatures, we therefore aim at learning control policies that are robust and adaptive to unknown, hidden contexts.

\paragraph{State space (\( \mathcal S \))}
Ideally, one could access the temporal profile of the pulse to describe the status of the laser system. Indeed, the temporal profile \( \chi(\psi) \) contains all the information needed to maximize peak intensity, including pulse energy and duration.
However, obtaining high-fidelity temporal profiles of ultra-short laser pulses in practice is a challenging task~\citep{trebino1993using, trebino1997measuring}.
Here, we instead leverage FROG traces as proxy for state information.
As FROG traces contain enough information to reconstruct temporal profiles~\citep{zahavy2018deep}, we argue they could also be used as direct inputs to a control policy aiming at maximizing peak intensity. Further, using FROG traces would be practically convenient given the availability of FROG detection devices in most HPL systems, and prevent the need for an intermediate step in the pulse shaping feedback loop to reconstruct \( \chi \) from its associated FROG trace, \( \Phi \). Hence, we directly include FROG traces \( \Phi_t \) in our state space. We complement states \( s_t \) with the vector of dispersion coefficients \( \psi_t \) and the action taken in the previous timestep, \( a_{t-1} \), giving \( s_t = \{ \Phi_t, \psi_t, a_{t-1} \} \), as they all are information available at test time.

\paragraph{Action space (\( \mathcal A \))}
As we are concerned with real world applicability of our method, we adopt an action space that is inherently machine-safe, and that can prevent erratically changing the control applied at test time~\citep{capuano2023temporl}. In this, we consider varying dispersion coefficients within predetermined boundaries defined at the level of the grated optical fiber, i.e. \( \psi_t \in [\psi_{\text{min}}, \psi_\text{max}]: c = \vert\psi_{\text{min}} - \psi_\text{max} \vert \). Actions are then defined as \( a_t \in [-\alpha c, +\alpha c] \), with \( \alpha \) being an arbitrary fraction of the total nominal range \( c \). In our method, we set \( \alpha=0.1 \), thus never changing \( \psi \) in one step by more than 10\% of the total possible variation.

\begin{figure}
    \centering
    \begin{minipage}{0.58\linewidth}
        \centering
        \includegraphics[width=\linewidth]{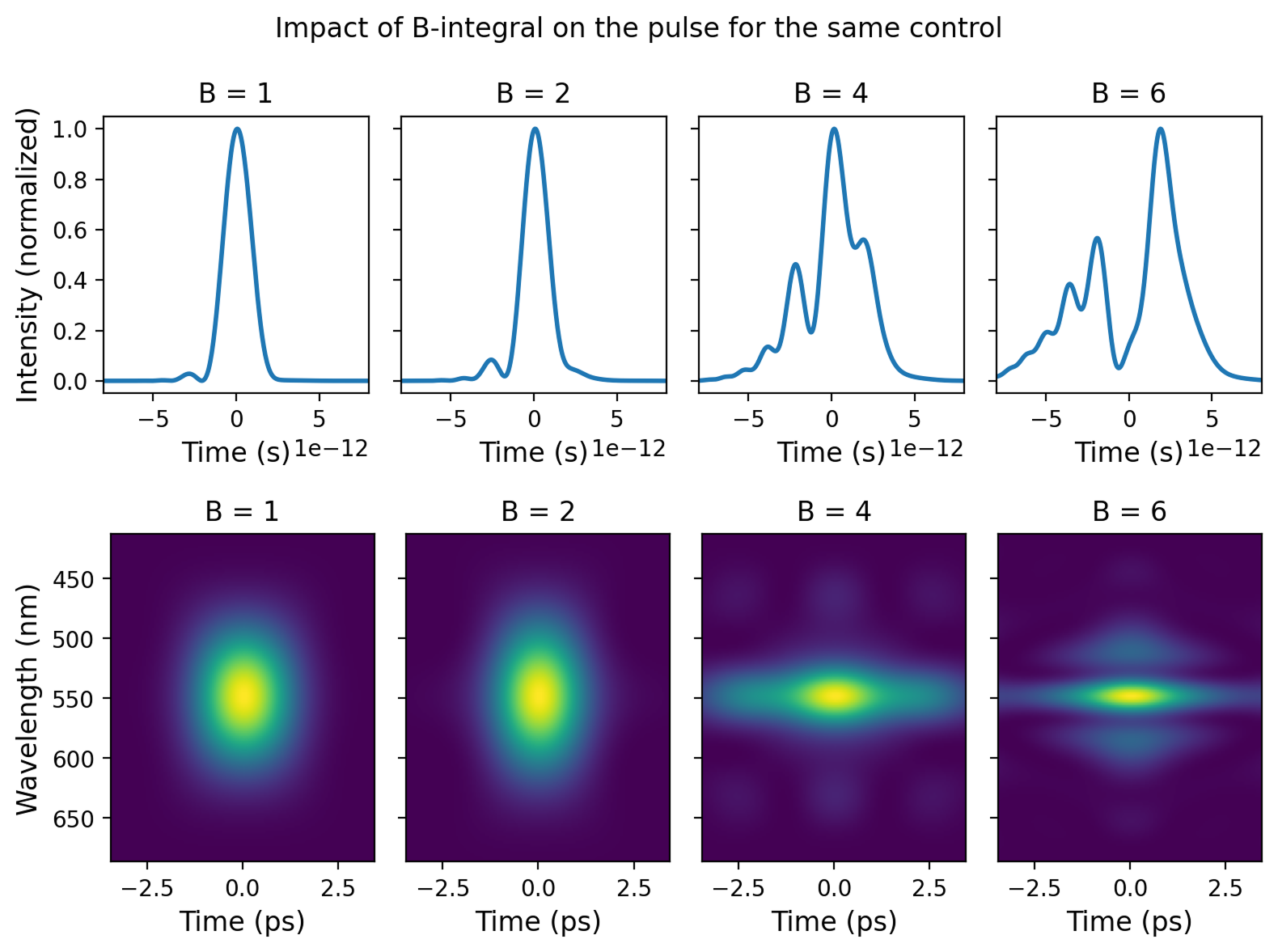}
        \caption{Impact of the B-integral parameter on the temporal profile (top) and FROG trace (bottom).}
        \label{fig:b_integral}
    \end{minipage}
    \hfill
    \begin{minipage}{0.4\linewidth}
        \centering
        \includegraphics[width=\linewidth]{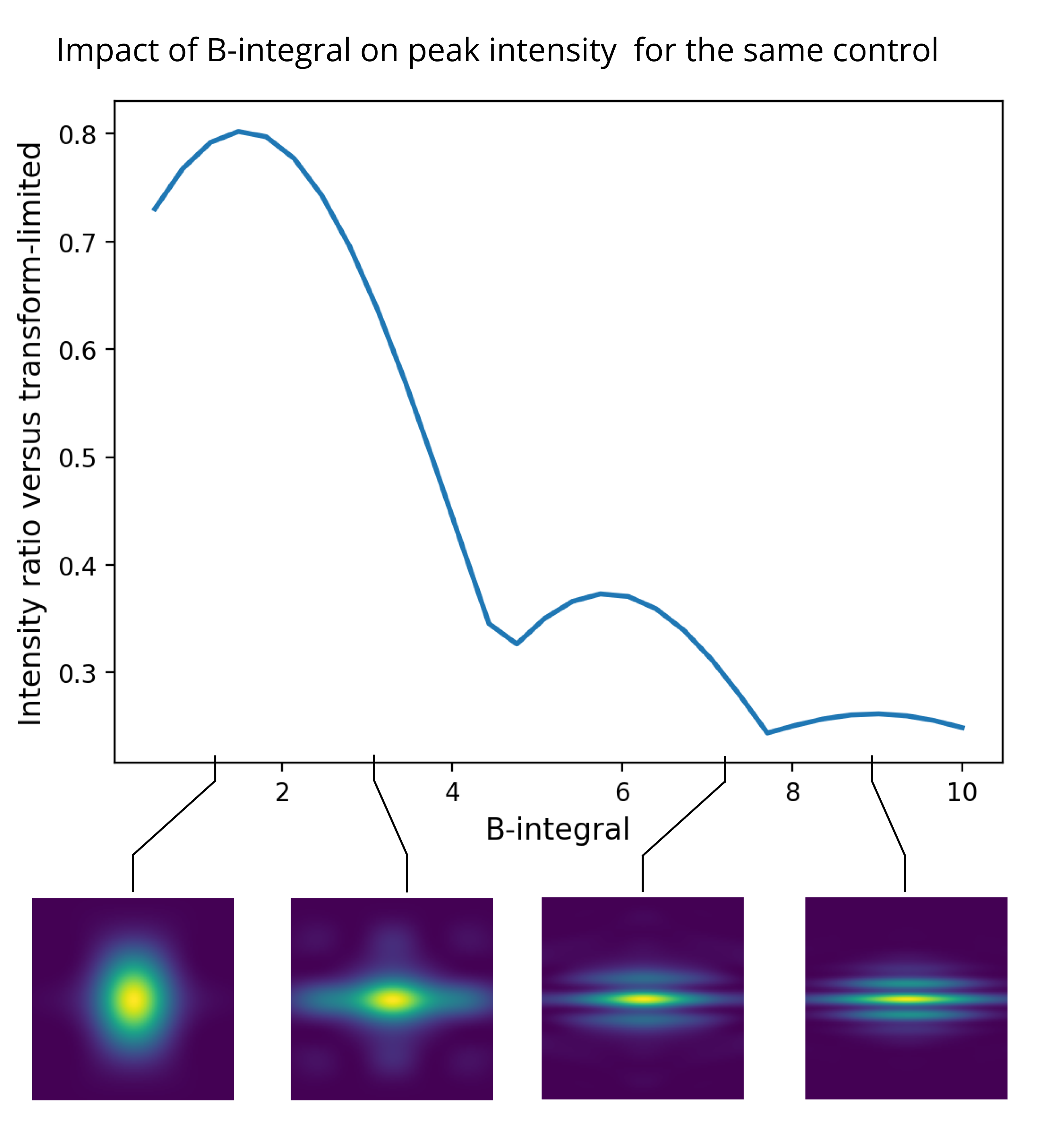}
        \caption{Impact of longer pulses on the peak intensity conveyed, measured as a fraction of $I_{TL}$.}
        \label{fig:dynamics_peak_intensity}
    \end{minipage}
\end{figure}

\paragraph{Environment dynamics ( \( \mathbb P_\xi: \mathcal S \times \mathcal A \times S \mapsto [0,1] \) ) }
Inspired by the successes of in-simulation learning in robotics~\citep{antonova2017reinforcement, akkaya2019solving, tiboni2023domain}, we employ simulations of the pump chain process while training a policy to control it. This allows us to scale the number of samples available at training time to amounts that are simply unfeasible on real-world laser hardware.
We refer the reader to~\cite{paschotta2008field} for an in-detail coverage of the physics describing the phase accumulation process, useful to model state-action-state transitions, \( \mathbb P_\xi (s_{t+1}\vert s_t, a_t) \). Here, we wish to pose particular emphasis on the role of \( \xi \) on \( \mathbb P_\xi \). Figure~\ref{fig:b_integral} shows how different \( \xi \)'s can lead to significantly different pulses when applying the same control \( \psi \). In particular, Figure~\ref{fig:b_integral} simulates the impact of randomizing the parameter regulating non-linear phase accumulation during amplification. This parameter is typically referred to as \emph{B-integral}, and indicated with \( B \). 
In HPL systems, one cannot typically assume to have control over \( B \) but indirectly: non-linear effects become more evident when higher-intensity pulses are propagated through non-linear crystal, which induces non-stationarity in \( B \). Further, precisely estimating \( B \) at a given time is a challenging tasks, prone to imprecision and which can have drastic impacts on the peak intensity achieved (Figure~\ref{fig:dynamics_peak_intensity}).

\paragraph{Reward function \( r \), starting condition \( \rho \) and discount factor \( \gamma \)}
We exploit our knowledge of HPL systems to design a reward function defined as the ratio between the current-pulse peak intensity \( I_t \) and the highest intensity possibly obtainable, \( I_{TL} \) achieved by so-called \emph{Transform-Limited} pulses, yielding \(r_t(s_t, a_t, s_{t+1}) = \frac{I_t}{I_{TL}} \in [0,1] \ \forall t \). In the absence of non-linear effects due to amplification, one would impose a phase on the stretcher that is opposite to the compressor's, \( \varphi_s(\omega) = - \varphi_c(\omega) \) so as to maximize intensity. As non-linearity is induced, it is reasonable to look for solutions in a neighborhood of the compressor's dispersion coefficients. Thus, one can use a multivariate normal distribution \( \mathcal N(-\psi_c, \epsilon \mathbb I) \) with mean \( -\psi_c \) and diagonal variance-covariance matrix. Lastly, we employed an episodic framework for this problem, fixing the number of total interactions to \( T=20 \), and used a discount factor of \( \gamma=0.9 \).

\subsection{Soft Actor Critic (SAC)}
Because we run training in simulation, we are able to drastically scale the experience available to the agent. With that being said, our simulation routine requires non-trivial computation, such as obtaining \( \Phi_t \) from \( \psi_t \). Thus, we limit ourselves to the generally more sample-efficient end of DRL, and refrain from using purely on-policy methods, such as TRPO/PPO~\citep{schulman2015trust, schulman2017proximal}.

SAC is an off-policy DRL algorithm learning Q-functions (policy \textit{evaluation}) that generalize across high-dimensional state-action spaces. Then, a stochastic policy is iteratively learned by explicitly maximizing the current Q-function estimate (policy \textit{improvement}). 
Interestingly, the Q-function itself is learned in a maximum entropy framework, leading to improved exploration and overall more effective learning over competing methods such as DDPG~\citep{haarnoja2018soft}. In this work, we implement both \emph{Vanilla} SAC and \emph{Asymmetric} SAC. The latter makes use of additional privileged information about the dynamics \( \xi \) while training. Notably, this information is yet not accessible by the policy, which is only conditioned on the current state.
The adoption of this asymmetric paradigm has proven empirically effective in easing the training process, by providing full information to the critic networks which are nevertheless not queried at test time~\citep{akkaya2019solving}.


\subsection{Domain Randomization (DR)}
To improve on the generalization of the control policy over unknown test conditions $\xi \sim \Xi^{real} $, we train a control policy in simulation by sampling dynamics parameters from an arbitrary auxiliary distribution $\Xi$. In particular, we compare two popular methods for choosing said distribution over $\xi$, namely Uniform Domain Randomization (UDR)~\citep{tobin2017domain, sadeghi2016cad2rl} and Domain Randomization via Entropy Maximization (DORAEMON)~\citep{tiboni2023domain}.

UDR models \( \Xi \) as a uniform distribution over manually defined bounds \( [\xi_{\min}, \xi_{\max}] \).
Crucially, identifying the bounds to use in training is an inherently brittle process: too-narrow bounds could hinder generalization, by not providing sufficient diversity over training. On the other hand, too-wide bounds can yield over-regularization, and thus result in reduced performance at test time. In the context of our application, experimentalists who are familiar with the specific pump-chain laser considered in this work estimate \( B \approx B_{\text{est}} = 2 \). Thus, we train a UDR policy in simulation by using $\xi = B \sim \mathcal U(1.5, 2.5)$, which is roughly equivalent to allowing misspecification of up to 25\% error.
However, even assuming access to ground-truth bounds, the probability mass of \( B \) is unlikely to be uniformly distributed on large supports---this would severely impact the performance of the system on a day to day basis. Conversely, it is reasonable to expect mass to be concentrated around some value within a possibly larger support, further away from \( B_{\text{est}} \).
In DORAEMON~\citep{tiboni2023domain}, the authors resolve the ambiguities in defining the training distribution by employing the principle of maximum entropy~\citep{jaynes1957information}. In other words, one could simply define a success indicator for the task, and seek for the maximum entropy training distribution \( \Xi \) that satisfies a lower bound on the success rate.

More precisely, DORAEMON solves this problem with a curriculum of evolving Beta distributions \( \Xi_k \sim \Beta(a_k, b_k) \).
In line with~\citet{tiboni2023domain}, we apply DORAEMON as an implicit meta-learning strategy for training adaptive policies over hidden dynamics parameters. We define a custom success indicator function on trajectories \( \tau_{\xi_k} \): terminal-state pulses \( \chi(\psi_T) \) must convey at least 65\% of the TL-intensity (\( I_T / I_{TL} \geq 0.65 \)) for the respective episode to be considered successful.
As a result, our implementation yields an automatic curriculum over DR distributions \( \Xi \) at training time such that entropy grows so long as the success rate is above \( 50 \)\%, as in the original DORAEMON paper.

\section{Experiments}
\label{sec:experiments}
\begin{figure}
    \centering
    \begin{subfigure}[b]{0.33\textwidth}
        \centering
        \includegraphics[width=\linewidth]{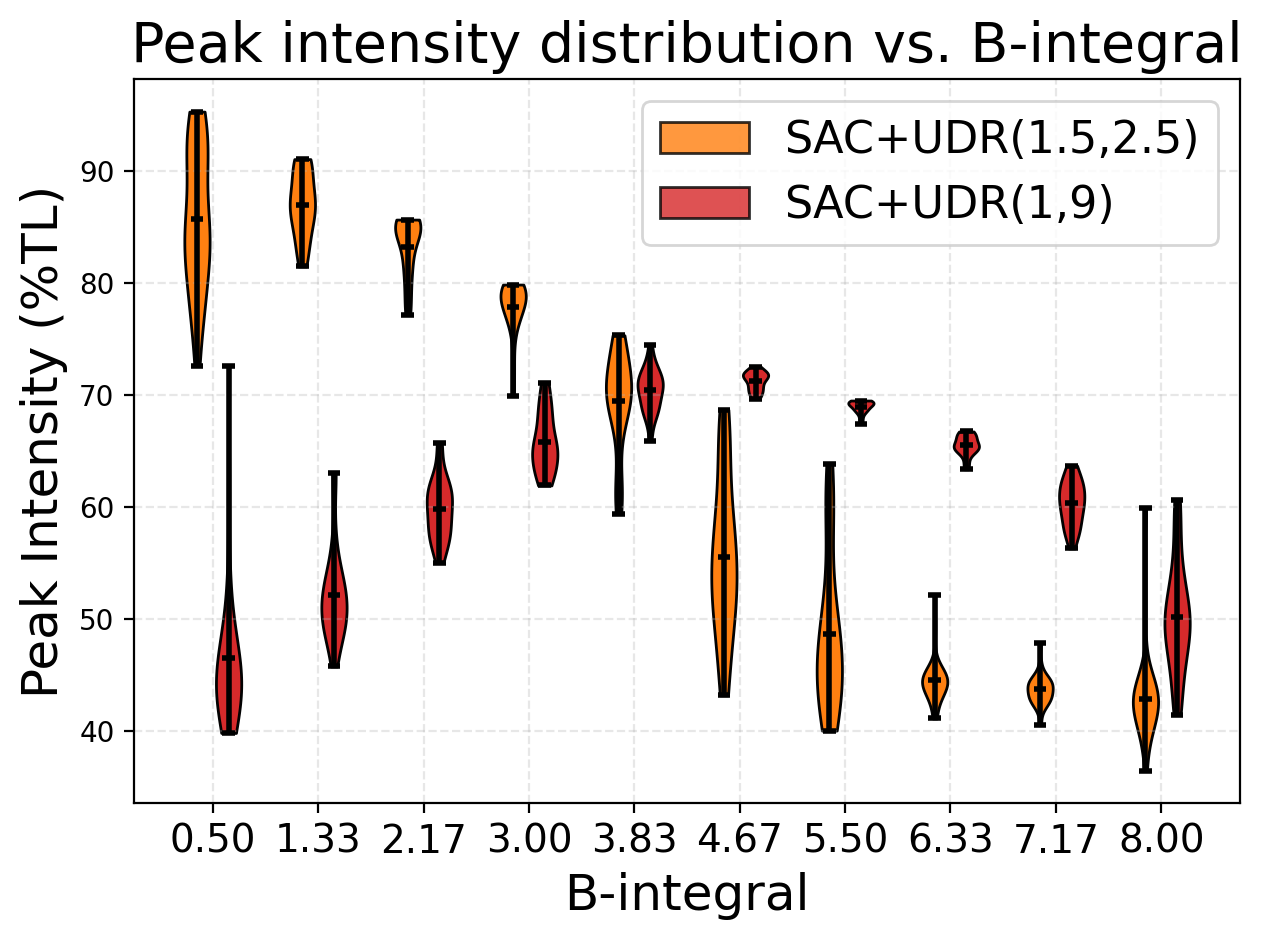}
        \caption{\( \mathcal{U}(1.5, 2.5), \ \mathcal{U}(1, 9) \)}
        \label{fig:picking_right_udr_hard}
    \end{subfigure}
    \hfill
    \begin{subfigure}[b]{0.33\textwidth}
        \centering
        \includegraphics[width=\linewidth]{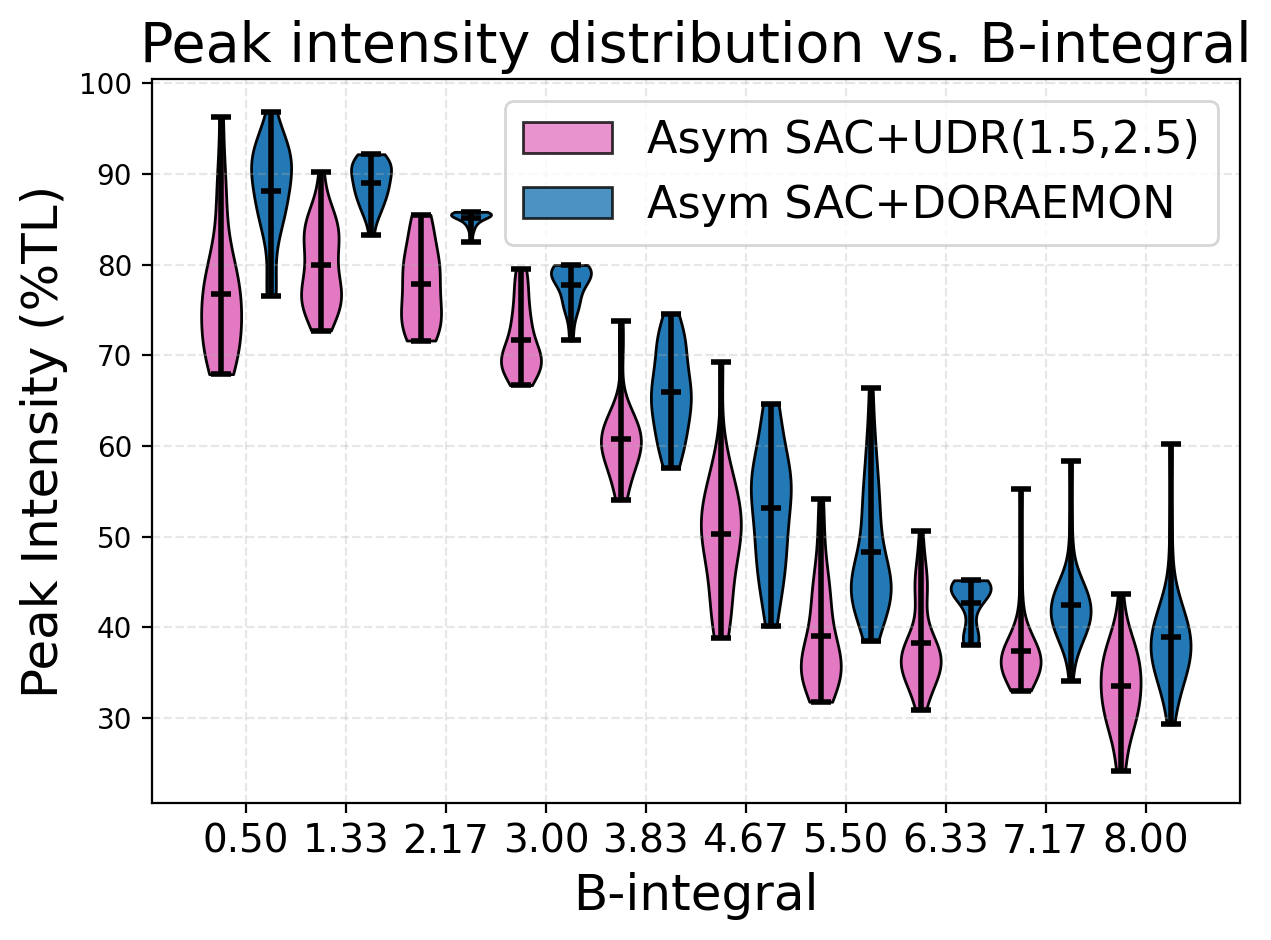}
        \caption{\( \mathcal{U}(1.5, 2.5) \), DORAEMON} 
        \label{fig:doraemon_outperforms_udr_violin}
    \end{subfigure}
    \hfill
    \begin{subfigure}[b]{0.32\textwidth}
        \centering
        \includegraphics[width=\linewidth]{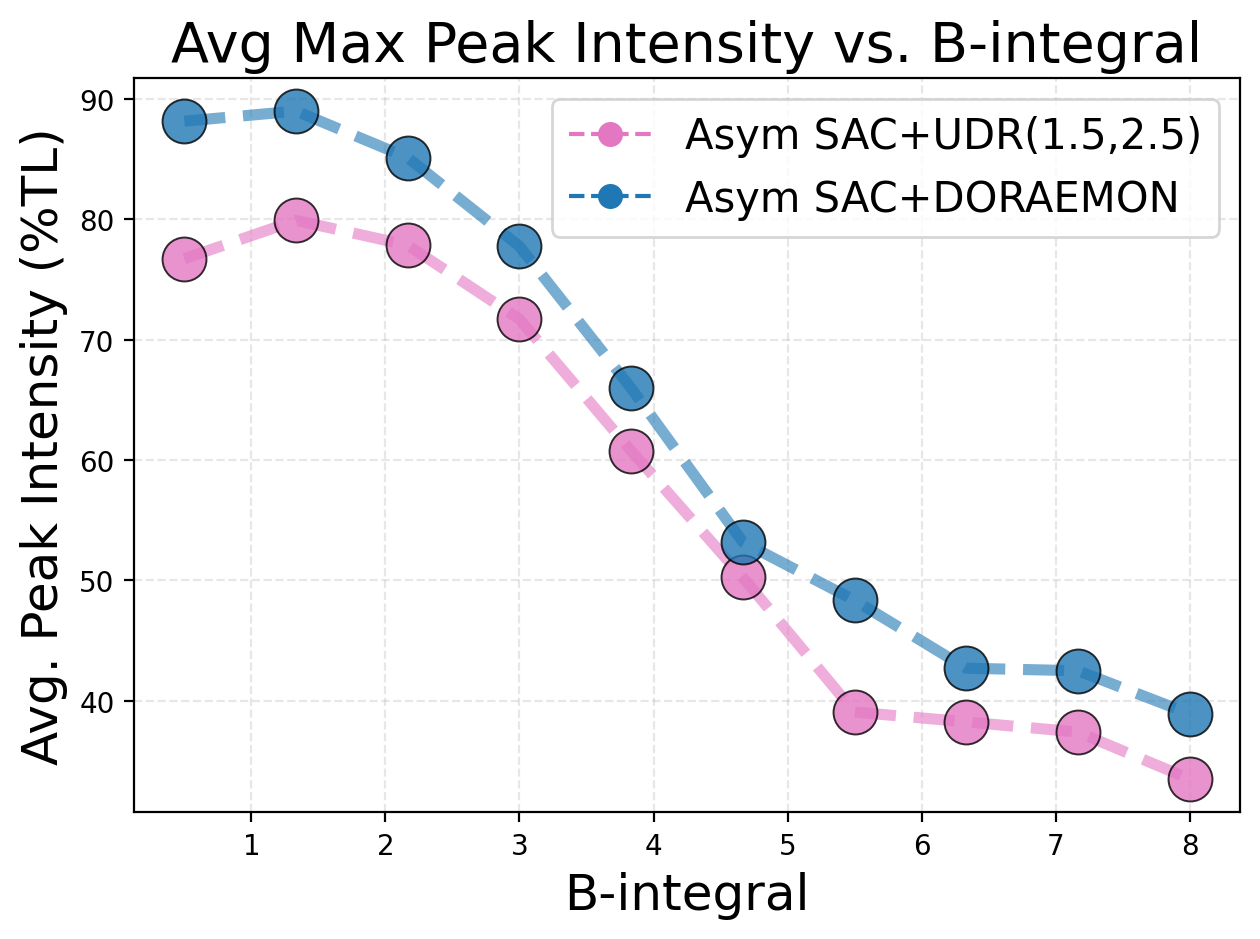}
        \caption{Average \( I^* \) versus \( B \)}
        \label{fig:doraemon_outperforms_udr_scatter}
    \end{subfigure}
    \caption{Distribution (a, b) of and average (c) of \( I^*/I_{TL} \) over 25 test episodes, for policies trained in different setup. Hand-picking the correct distribution for DR is an error prone process (a). Automated approaches like DORAEMON offer a promising alternative adapting the DR distribution using signal from training (b), resulting in an overall better performance over hand-picked distributions across diverse values of \( B \) (c).}
    \label{fig:max_intensity_vs_b_integral}
\end{figure}

\begin{table}
    \centering
    \caption{Success rate over 25 test episodes: proportion of episodes with \( I^* / I_{TL} \geq 70\% \) in multiple experimental conditions}
    \label{tab:succ_rate}
    {%
    \resizebox{0.9\textwidth}{!}{%
    \begin{tabular}{cccccc}
        \textbf{Algorithm} & \textbf{\begin{tabular}[c]{@{}c@{}}Training\\ timesteps\end{tabular}} & \textbf{\begin{tabular}[c]{@{}c@{}}Training\\ Distribution\end{tabular}} & \textbf{\begin{tabular}[c]{@{}c@{}}Success Rate \\ (\(B=0.5\))\end{tabular}} & \textbf{\begin{tabular}[c]{@{}c@{}}Success Rate \\ (\(B=2.17\))\end{tabular}} & \textbf{\begin{tabular}[c]{@{}c@{}}Success Rate \\ (\(B=3.83\))\end{tabular}} \\ \hline
        SAC                & 200k                                                                  & \( \mathcal{U}(1.5, 2.5) \)                                              & 1.00                                                                         & 1.00                                                                          & 0.60                                                                          \\
        \rowcolor[HTML]{EFEFEF} 
        SAC                & 200k                                                                  & \( \mathcal{U}(1, 9) \)                                                  & 0.04                                                                         & 0.00                                                                          & 0.64                                                                          \\
        Asymmetric-SAC     & 200k                                                                  & \( \mathcal{U}(1.5, 2.5) \)                                              & 0.84                                                                         & 1.00                                                                          & 0.08                                                                          \\
        \rowcolor[HTML]{EFEFEF} 
        Asymmetric-SAC     & 200k                                                                  & DORAEMON(1, 3.5)                                                         & 1.00                                                                         & 1.00                                                                          & 0.28                                                                         
    \end{tabular}
    }
    }
\end{table}

\begin{figure}
        \centering
        \begin{minipage}{0.45\textwidth}
            \centering
            \includegraphics[width=\linewidth]{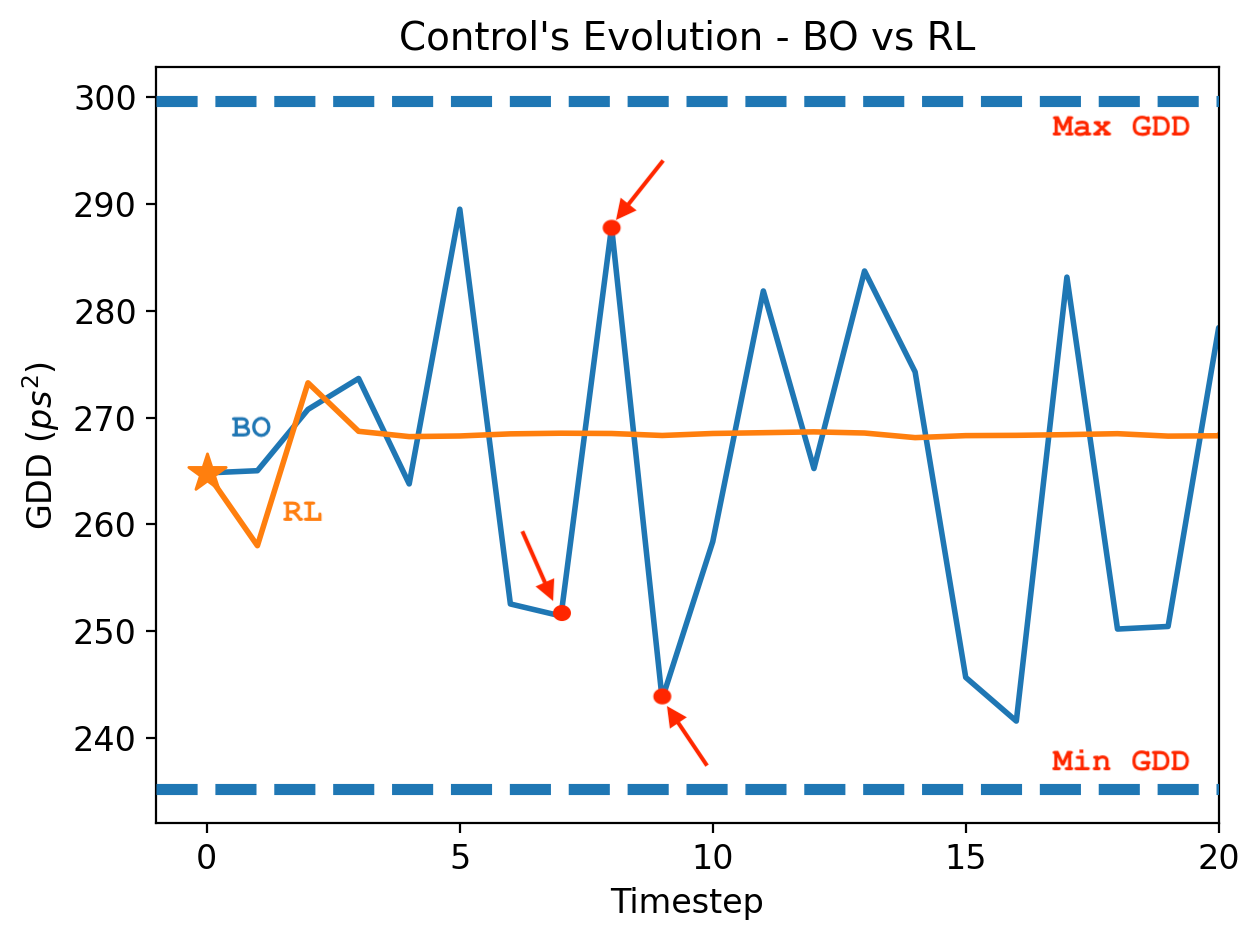}
            \caption{Controls applied (BO vs RL). As it samples from an iteratively-refined surrogate model of \(I(\psi)\), BO explores much more erratically than RL.}
            \label{fig:bayes_vs_rl}
        \end{minipage}
        \hfill
        \begin{minipage}{0.45\textwidth}
            \centering
            \includegraphics[width=\linewidth]{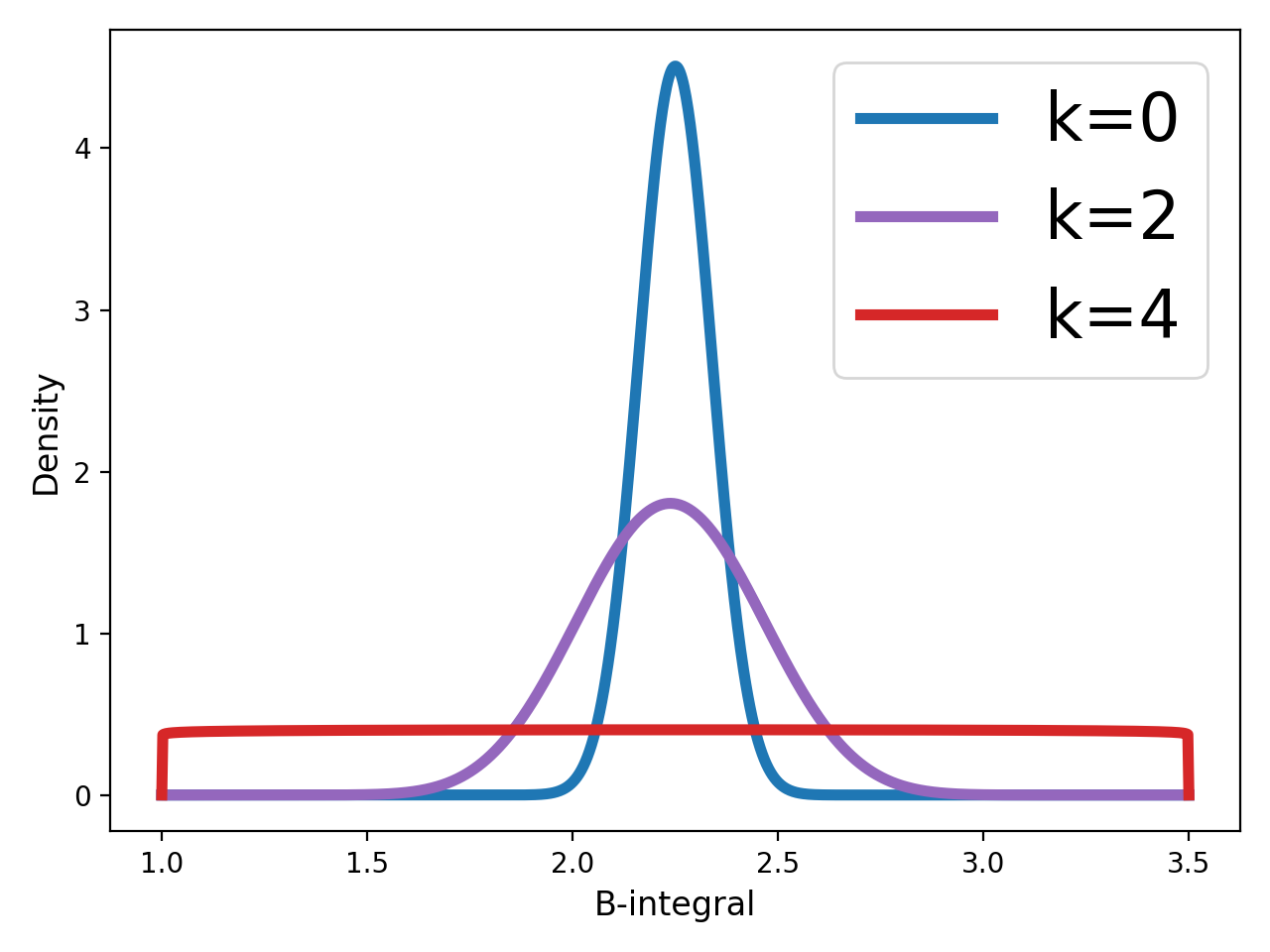}
            \caption{Evolution of the distribution used when training an agent with DORAEMON. Further updates \( k \in [4,20] \) do not impact the evolution of the distribution over \( B \).}
            \label{fig:DORAEMON_distrs_over_training}
        \end{minipage}
        \label{fig:control_and_distribution}
\end{figure}

We validate our claims on the improved machine-safety of RL over popular baselines such as BO~\citep{shalloo2020automation} by comparing the evolution of the controls applied at test time for both BO and \emph{mini-SAC}. 
As BO cannot be used to process images, we benchmark it against \emph{mini-SAC}, a simplified version of our algorithm that exclusively uses \( \psi \) in the state vector.
We report the comparison between mini-SAC and BO to further show the benefits in terms of machine safety of DRL.
Figure~\ref{fig:bayes_vs_rl} displays the evolution of the controls applied over the first 20 interactions between BO and the RL-based controller. 
Unlike BO's solutions, which are stationary and can only be transferred assuming high-fidelity simulations, RL policies can be transferred across domains.
Notably, this allows us to allocate dangerous erratic exploration to in-simulation training, preventing erratic behavior at test time---similarly to established work in robotics~\citep{kober2013reinforcement}.

\begin{figure}
    \centering
    \includegraphics[width=\linewidth]{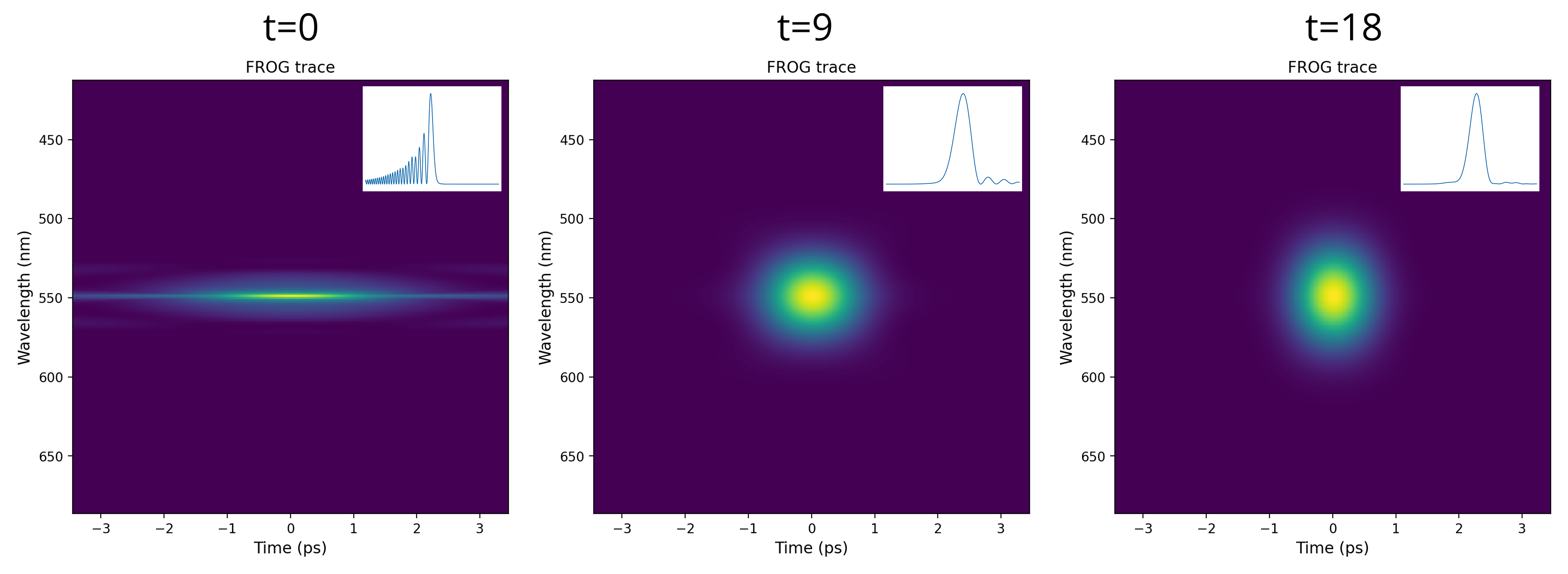}
    \caption{SAC, learning to shape temporal pulses directly from FROG traces. The temporal profile associated with the FROG trace is superimposed on the top right, and is never made available to the agent. In under 20 interactions, the agent produces near-TL pulses.}
    \label{fig:frog_opt}
\end{figure}

Since temporal profiles \( \chi(\psi) \) are typically unavailable, we exclusively rely on 64x64 single-channel images as state representations for the agent, as discussed in~\cref{sec:mdp}.
Table~\ref{tab:peak_intensities} shows the average max peak intensity over 25 test episodes, after training SAC for 200k timesteps, while Figure~\ref{fig:frog_opt} shows the FROG traces obtained during a test episode at various timesteps. Crucially, our policy does learn to control \( \psi \) to compress the pulse in time, achieving an average of 85.12\% of TL's peak intensity around \(B_{\text{est}} \), with peaks close to 90\% for oscillations in \( B \) (Table~\ref{tab:max_peak_intensities}). These findings also attest the effectiveness of using single-channel images as affordable proxy input to maximize peak intensity.

\begin{table}
    \centering
    \caption{Average (plus-minus standard deviation) maximal peak intensity over 25 test episodes, for a combination of algorithms, training and testing conditions. We test our algorithms on fixed values of \( B \).
    }
    \label{tab:peak_intensities}
    \resizebox{\textwidth}{!}{%
    \begin{tabular}{cccccc}
        \textbf{Algorithm} & \textbf{\begin{tabular}[c]{@{}c@{}}Training\\ timesteps\end{tabular}} & \textbf{\begin{tabular}[c]{@{}c@{}}Training\\ Distribution\end{tabular}} & \textbf{\begin{tabular}[c]{@{}c@{}}Avg. Max Peak \\ Intensity (\(B=0.5\))\end{tabular}} & \textbf{\begin{tabular}[c]{@{}c@{}}Avg. Max Peak \\ Intensity (\(B=2.17\))\end{tabular}} & \textbf{\begin{tabular}[c]{@{}c@{}}Avg. Max Peak \\ Intensity (\(B=3.83\))\end{tabular}} \\ \hline
        SAC                & 200k                                                                  & \( \mathcal{U}(1.5, 2.5) \)                                              & 85.76 ± 6.19                                                                            & 83.23 ± 2.66                                                                             & 69.49 ± 4.35                                                                             \\
        \rowcolor[HTML]{EFEFEF} 
        SAC                & 200k                                                                  & \( \mathcal{U}(1, 9) \)                                                  & 46.51 ± 6.83                                                                            & 59.87 ± 2.66                                                                             & 70.43 ± 1.87                                                                             \\
        Asymmetric-SAC     & 200k                                                                  & \( \mathcal{U}(1.5, 2.5) \)                                              & 76.73 ± 6.90                                                                            & 77.86 ± 4.38                                                                             & 60.77 ± 4.23                                                                             \\
        \rowcolor[HTML]{EFEFEF} 
        Asymmetric-SAC     & 200k                                                                  & DORAEMON(1, 3.5)                                                         & 88.16 ± 5.22                                                                            & 85.12 ± 0.77                                                                             & 65.98 ± 4.70                                                                            
    \end{tabular}
    }
\end{table}

\begin{figure}
    \centering
    \begin{subfigure}{0.32\textwidth}
        \centering
        \includegraphics[width=\linewidth]{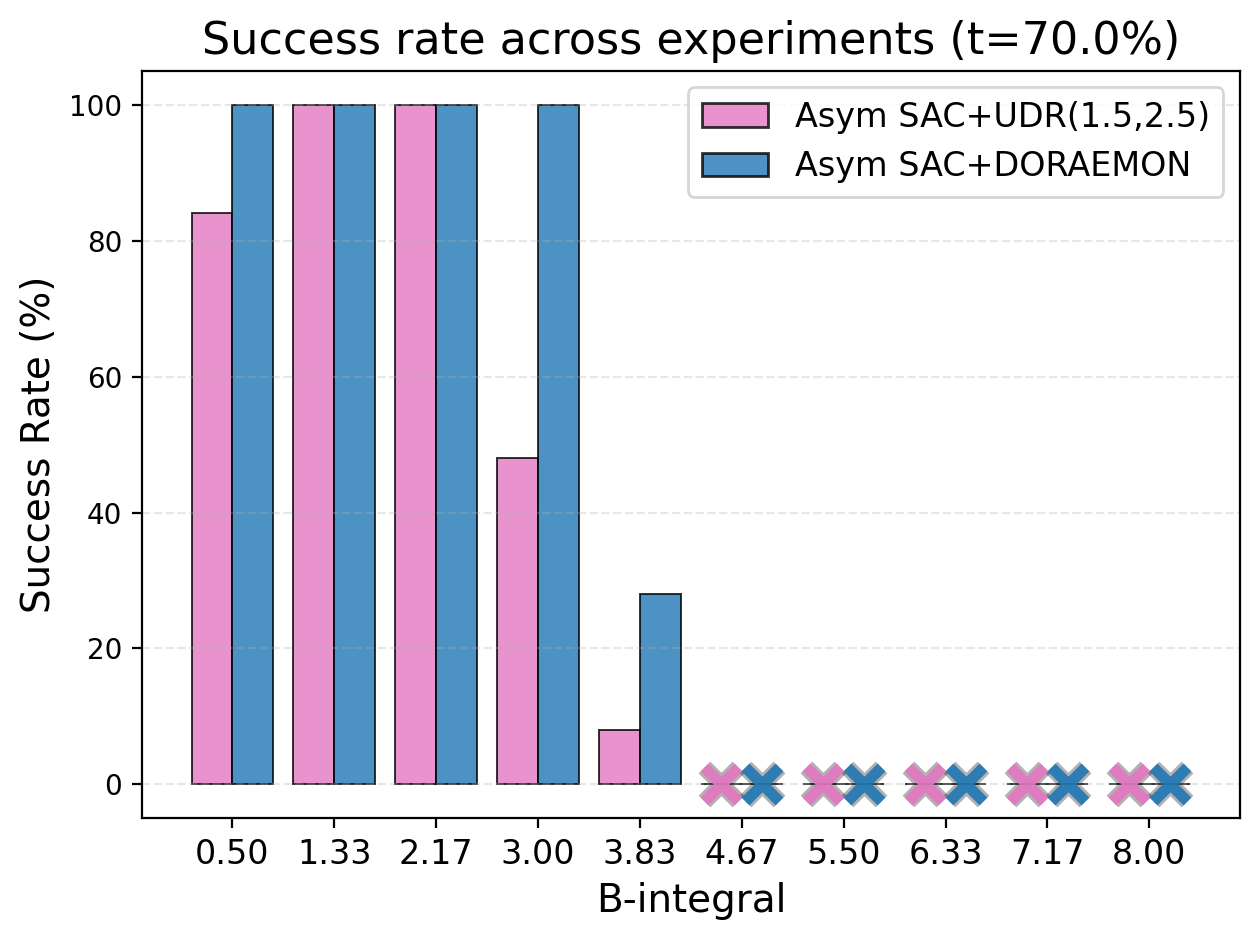}
        \caption{ \( t = 70 \% \) }
        \label{fig:succ_rate_70}
    \end{subfigure}
    \hfill
    \begin{subfigure}{0.32\textwidth}
        \centering
        \includegraphics[width=\linewidth]{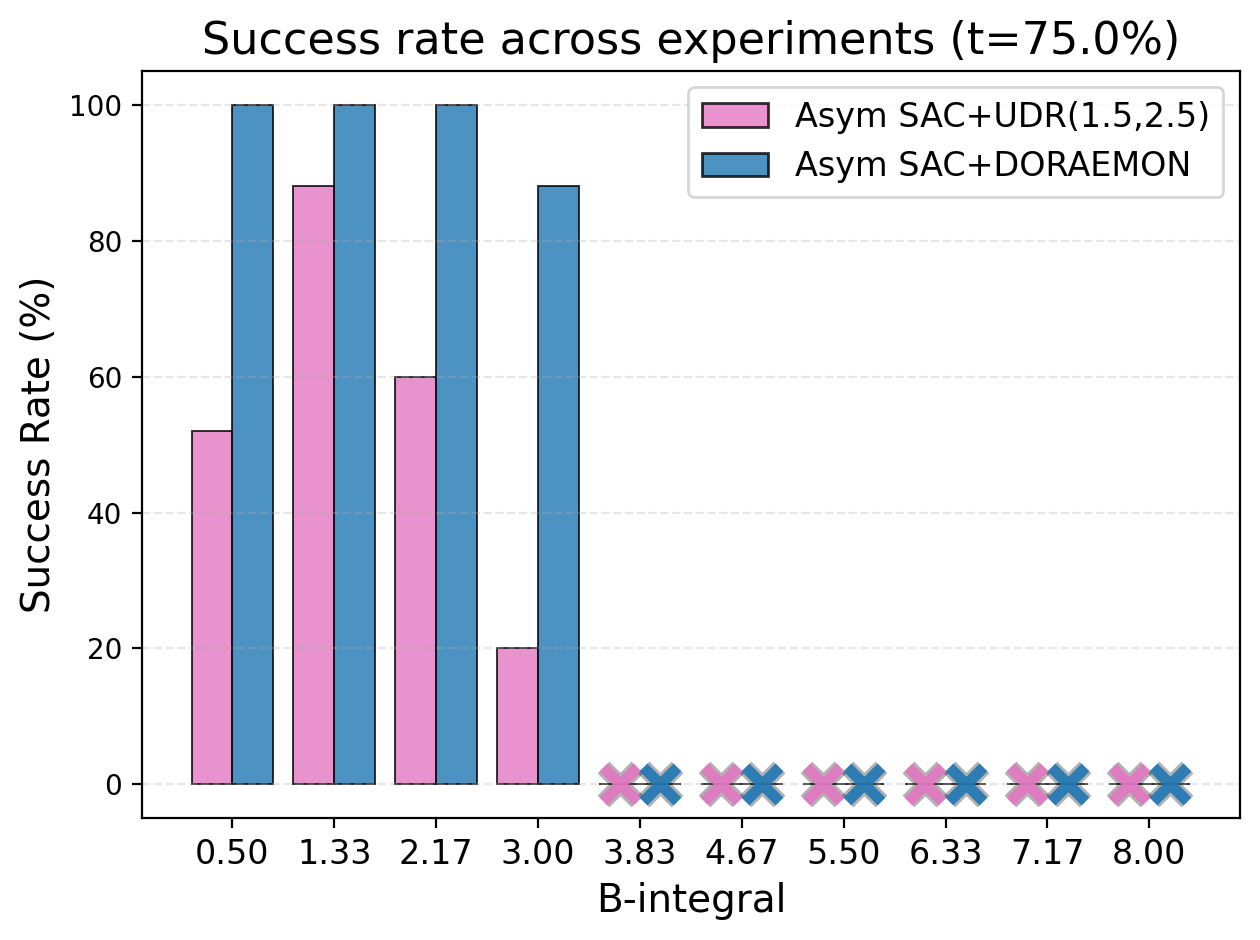}
        \caption{\( t = 75 \% \)} 
        \label{fig:succ_rate_75}
    \end{subfigure}
    \hfill
    \begin{subfigure}{0.32\textwidth}
        \centering
        \includegraphics[width=\linewidth]{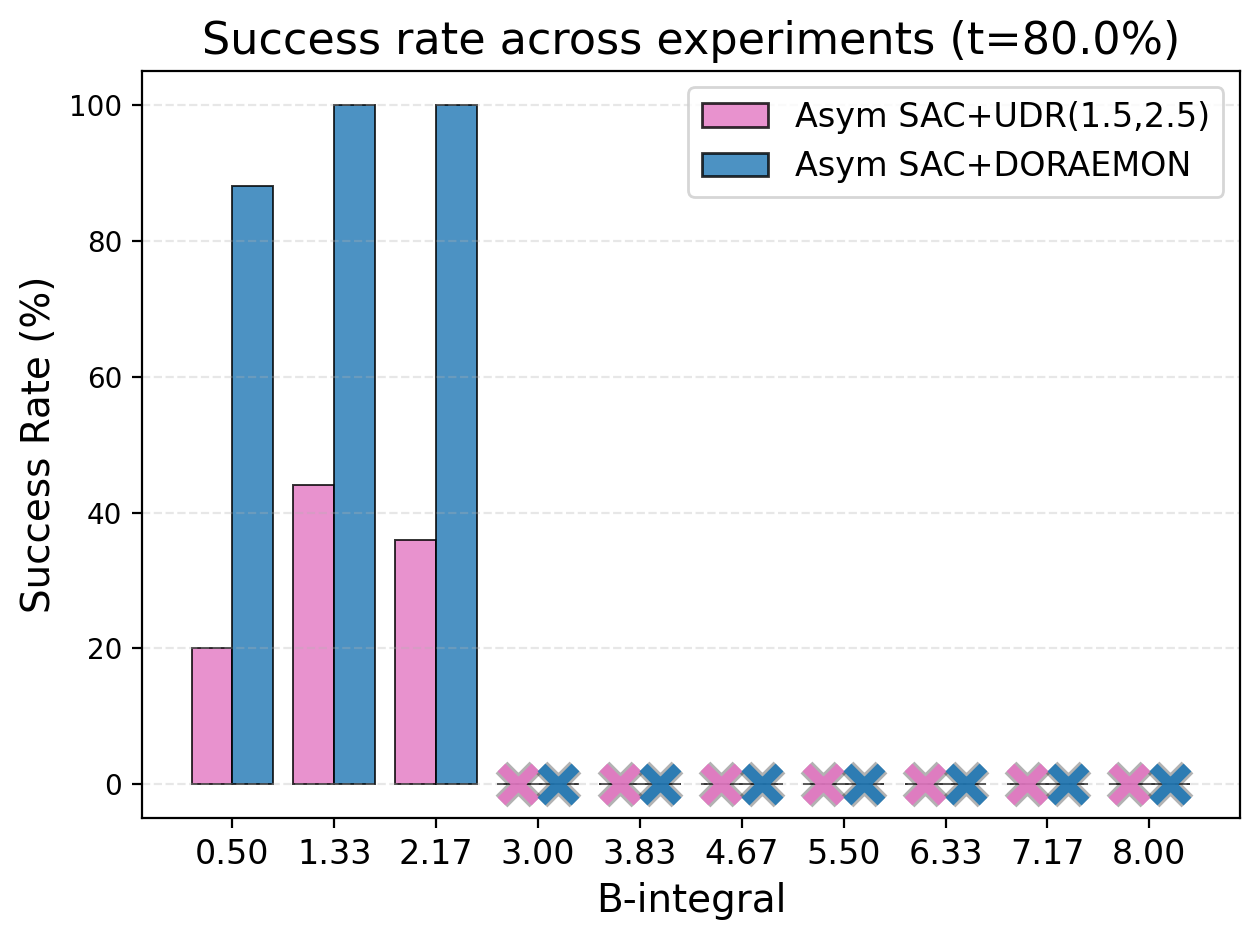}
        \caption{\( t = 80 \% \)}
        \label{fig:succ_rate_80}
    \end{subfigure}
    \caption{Success rates for DORAEMON vs \( \mathcal U(1.5, 2.5) \) across different thresholds. Bars indicates the percentage of the 25 test episodes achieving peak intensities above the specified threshold value. Crosses indicate no test episode achieved \(I^*/I_{TL} \geq t \).}
    \label{fig:succ_rate}
\end{figure}

Later, we benchmark the robustness of our policy to changes in the environment dynamics.
Particularly, we employ DR during training, and use Asymmetric-SAC together with a stack of the last \( n=5 \) states. Using the resulting \emph{history-based} policy holds promise in the context of DR to promote adaptive, meta-learning behavior~\citep{chen2021understanding, tiboni2023domain, akkaya2019solving}.
We evaluate the performance of our method by measuring the average max intensity versus equally-spaced changes in the value of B-integral (Figure~\ref{fig:max_intensity_vs_b_integral}). We then zoom in on the values believed to represent the system's state more realistically, and report in Table~\ref{tab:peak_intensities} the average peak intensity, alongside the standard deviation. Further, we report in Table~\ref{tab:max_peak_intensities} the minimum and maximum peak intensity measured over 25 test episodes.

We observe the performace of policies to significantly vary based on the distribution used while training (Figure~\ref{fig:picking_right_udr_hard}, Figure~\ref{fig:doraemon_outperforms_udr_violin}), motivating the adoption of an automated DR method such as DORAEMON (Figure~\ref{fig:doraemon_outperforms_udr_scatter}).
Table~\ref{tab:peak_intensities} shows the impact of choosing narrower rather than wider bounds for UDR, as we find wider UDR to cause over-regularization, hindering performance.
We compare a naive UDR approach---using realistic bounds suggested by human experts familiar with the system---with DORAEMON, by adapting the training distribution \( \{ \Xi_{k} \}_{k=1}^K\) across \( K=20 \) steps over 200k timesteps. Compared to UDR, DORAEMON displays better test-time performance around our estimate \( B_{\text{est}} = 2 \), and generally provides superior success rate (Table~\ref{tab:succ_rate}).

Figure~\ref{fig:DORAEMON_distrs_over_training} shows the evolution of the distributions generated by DORAEMON \(\{\Beta(a_k, b_k) \}_{k=1}^{K} \) over the course of training. Interestingly, distributions eventually converge to the maximum entropy \( \mathcal U(1, 3.5) \), indicating that sufficient training performance can be maintained even in the extreme case. To investigate the effectiveness of the curriculum for DORAEMON, we then evaluate it against a naive UDR baseline using \( \mathcal U(1.5, 2.5) \)---a range indicated as realistic by domain experts---and observe DORAEMON's superiority in both Figure~\ref{fig:doraemon_outperforms_udr_scatter} and Table~\ref{tab:peak_intensities}.

Lastly, we report the success rates using an indicator function over the peak intensity achieved in a rollout \( I^* \), \(f_t(I^*) = \mathbf{1}_{I^* \geq t} \). We first present success rates for different threshold value \( t \in [70\%, 75\%, 80\%] \) in Figure~\ref{fig:succ_rate} and then for \( t = 70\% \) only in Table~\ref{tab:succ_rate}.
DORAEMON consistently outperforms \( \mathcal{U}(1.5, 2.5) \), underscoring the premise of the method in our context. Further, DORAEMON exhibits stronger performance in settings affected by less evident non-linear effects (\( B \leq 2.5 \)), more often encountered in practice.

\begin{table}
    \centering
    \caption{Min-Max ranges for the maximal peak intensity over 25 test episodes, for a combination of algorithms, training and testing conditions.
    }
    \label{tab:max_peak_intensities}
    \resizebox{\textwidth}{!}{%
    \begin{tabular}{cccccc}
        \textbf{Algorithm} & \textbf{\begin{tabular}[c]{@{}c@{}}Training\\ timesteps\end{tabular}} & \textbf{\begin{tabular}[c]{@{}c@{}}Training\\ Distribution\end{tabular}} & \textbf{\begin{tabular}[c]{@{}c@{}}Min/Max Peak \\ Intensity (\(B=0.5\))\end{tabular}} & \textbf{\begin{tabular}[c]{@{}c@{}}Min/Max Peak \\ Intensity (\(B=2.17\))\end{tabular}} & \textbf{\begin{tabular}[c]{@{}c@{}}Min/Max Peak \\ Intensity (\(B=3.83\))\end{tabular}} \\ \hline
        SAC                & 200k                                                                  & \( \mathcal{U}(1.5, 2.5) \)                                              & 72.63/95.31                                                                            & 77.19/85.67                                                                             & 59.39/75.34                                                                             \\
        \rowcolor[HTML]{EFEFEF} 
        SAC                & 200k                                                                  & \( \mathcal{U}(1, 9) \)                                                  & 39.84/72.60                                                                            & 55.03/65.74                                                                             & 65.90/74.48                                                                             \\
        Asymmetric-SAC     & 200k                                                                  & \( \mathcal{U}(1.5, 2.5) \)                                              & 67.92/96.30                                                                            & 71.62/85.49                                                                             & 54.05/73.75                                                                             \\
        \rowcolor[HTML]{EFEFEF} 
        Asymmetric-SAC     & 200k                                                                  & DORAEMON(1, 3.5)                                                         & 76.57/96.86                                                                            & 82.53/85.81                                                                             & 57.55/74.57                                                                            
    \end{tabular}%
    }
\end{table}

\section{Conclusions}
\label{sec:discussion}
In this work, we present a novel application of RL to the rich and complex domain of experimental laser physics, using RL as the backbone for a fully automated pulse-shaping routine. Leveraging domain knowledge of the processes regulating phase accumulation in HPL systems, we design a coarse simulator of the pump chain of a HPL system, which we use to develop control strategies that exclusively use non-destructive measurements in the form of images to maximize the peak intensity of ultra-short laser pulses.

We compare our controller under diverse experimental conditions in simulation, and observe markedly gentler exploration and peak intensities of up to $90\%$ of the transform-limited reference. In addition, we reformulate pulse shaping as a Latent MDP and leverage recent domain randomisation techniques to obtain policies that maintain performance under moderate variations of the dynamics parameters.

\paragraph{Limitations}
We identify several limitations remaining in our contribution. In particular, HPL systems' performance is known to be influenced, alongside B-integral, by the dispersion coefficients of the compressor. These dispersion coefficients are highly sensitive to the delicate alignment of the compressor optics, which is typically a cumbersome and time-consuming process in ultra-fast optics. As such, we concluded randomizing over these coefficients was unnecessary in a first instance, as a great deal of effort and diagnostic is spent in properly assessing and monitoring the compressor's status. Still, adapting to their variation as well is a very promising approach, which we seek to investigate further. Moreover, a second limitation is the sample inefficiency of our method, requiring hundreds of thousands to samples to discover well performing policies. We argue this is particularly problematic considering the amount of knowledge available on the process of phase accumulation in linear and non-linear crystals. While our coarse simulator provides a useful tool for model-free learning, the absence of explicit modeling of the dynamics limits data efficiency. Integrating model-based components could significantly improve sample efficiency.

Despite these limitations, our work takes a significant step toward the integration of DRL in HPL systems, providing a framework that is both practical and adaptable to experimental constraints, and prove the effectiveness of the technique in ultra-short laser physics.





\bibliography{main}
\bibliographystyle{rlj}


\end{document}